\documentclass[letterpaper, 10 pt, conference]{ieeeconf}  % Comment this line out
\IEEEoverridecommandlockouts% This command is only needed if you want to use the \thanks command
\usepackage{graphics} % for pdf, bitmapped graphics files
\usepackage{epsfig} % for postscript graphics files
\usepackage{mathptmx} % assumes new font selection scheme installed
\usepackage{times} % assumes new font selection scheme installed
\usepackage{amsmath} % assumes amsmath package installed
\usepackage{amssymb}  % assumes amsmath package installed
\usepackage{booktabs}  
\usepackage{textcomp}
\usepackage{gensymb}
\title{\LARGE \bf Mesh-based 3D Textured Urban Mapping}

\author{Andrea Romanoni$^{1}$  Daniele Fiorenti$^{1}$ Matteo Matteucci$^{1}$
\thanks{$^{1}${Politecnico di Milano, Dipartimento di Elettronica, Informazione e Bioingegneria (DEIB),  Milano, Italy}
        \vfill{\tt\scriptsize andrea.romanoni@polimi.it (corresponding author)}
		\vfill{\tt\scriptsize daniele.fiorenti@mail.polimi.it} 
        \vfill{\tt\scriptsize matteo.matteucci@polimi.it}
        }
        }%

\begin{document}
\maketitle
\thispagestyle{empty}
\pagestyle{empty}

%%%%%%%%%%%%%%%%%%%%%%%%%%%%%%%%%%%%%%%%%%%%%%%%%%%%%%%%%%%%%%%%%%%%%%%%%%%%%%%%
\begin{abstract}
In the era of autonomous driving, urban mapping represents a core step to let vehicles interact with the urban context. 
Successful mapping algorithms have been proposed in the last decade building the map leveraging on data from a single sensor.
The focus of the system presented in this paper is twofold: the joint estimation of a 3D map from lidar data and images, based on a 3D mesh, and its texturing.
Indeed, even if most surveying vehicles for mapping are endowed by cameras and lidar, existing mapping algorithms usually rely on either images or lidar data; moreover both image-based and lidar-based systems often represent the map as a point cloud, while a continuous textured mesh representation would be useful for visualization and navigation purposes.
%In this paper, we propose a novel mapping framework which reconstructs a textured mesh of the environment by relying both on lidar and image data: 
In the proposed framework, we join the accuracy of the 3D lidar data, and the dense information and appearance carried by the images, in estimating a visibility consistent map upon the lidar measurements, and refining it photometrically through the acquired images. 
%In the proposed framework we need to explicitly detect and remove moving objects in the laser point cloud with a state-of-the-art algorithm \cite{postica16}, and we exploit this output to filter properly the moving object in the image-based photometric refinement and in the texturing.
%Moreover, since visibility consistency applied to laser beams that traverse the car windows causes unwanted holes, we also propose an effective and efficient method to detect the cars in the lidar point cloud, and we explicitly model and add them to the map.
We evaluate the proposed framework against the KITTI dataset and we show the performance improvement with respect to two state of the art urban mapping algorithms, and two widely used surface reconstruction algorithms in Computer Graphics.
\end{abstract}

% \begin{IEEEkeywords}
% Mapping, Autonomous Vehicle Navigation,Computer Vision for Other Robotic Applications
% \end{IEEEkeywords}
%%%%%%%%%%%%%%%%%%%%%%%%%%%%%%%%%%%%%%%%%%%%%%%%%%%%%%%%%%%%%%%%%%%%%%%%%%%%%%%%
\section{Introduction}
The growing interests around autonomous driving has focused the Robotics and Computer Vision communities on specific research areas such as sensing, mapping, and driving policy development~\cite{shashua2016autonomous}. 
Here we focus on urban dense textured mapping, which plays an important role to enable autonomous navigation through cities.

Surveying vehicles, aiming at city mapping, are usually endowed with 360\degree\  laser range finders and monocular or stereo cameras. 
While laser range finders directly provide accurate 3D measurements of the environment, cameras collect its dense appearance. 
Successful mapping algorithms have been proposed using cameras or lasers separately, but, to the best of our knowledge, none of them exploits both data sources to build dense textured maps. 
Only~\cite{maddern2016real} considers both information sources, but the final outcome is a dense stereo matching disparity map, and not a full photoconsistent 3D map of the environment.
Moreover, most systems reconstruct a point cloud or a voxelized map, while a continuous mesh of the environment which represents in a dense way the observed scene can lead to  more robust navigation or localization and it  allows texturing too. 

Laser-based mapping algorithms~\cite{hornung2013octomap} and~\cite{khan2015adaptive} have shown to produce accurate maps of the environment, but fine-grained details are often discarded, due to sparsity of the 3D point clouds, and they can include moving points if not properly filtered from the laser scans.
Feature-based image-based mapping algorithms~\cite{litvinov_lhiuller14}, ~\cite{romanoni15b} and ~\cite{wu13} build the map reconstruction on 3D points estimated through robust 2D to 2D correspondences; they are able to discard moving points from the final outcome, nevertheless the resulting map is still a sparse point cloud.
Finally, photometric image-based approaches from multi-view stereo exploit the whole information carried in the image and result in more detailed dense or semi-dense reconstructions; however they usually assume the scene to be static not being designed to cope with moving objects.

\begin{figure}[tp]
 \centering
    \begin{tabular}{c}
	\hspace{-0.3cm}\includegraphics[width=0.5\textwidth]{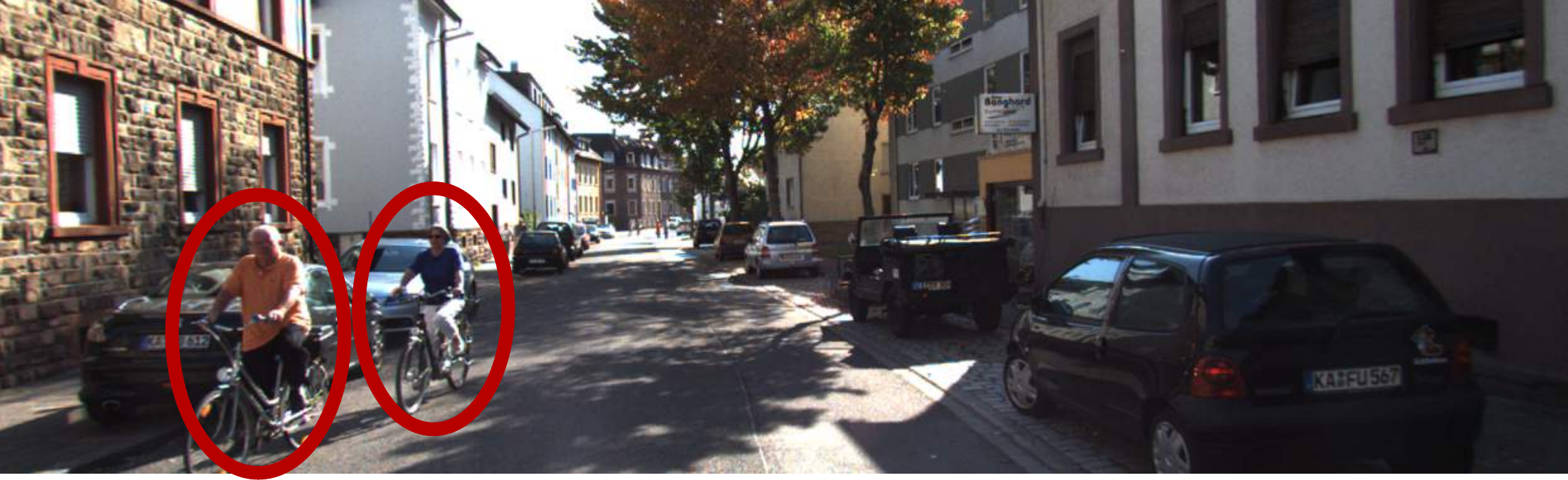}\\
	\hspace{-0.3cm}\includegraphics[width=0.5\textwidth]{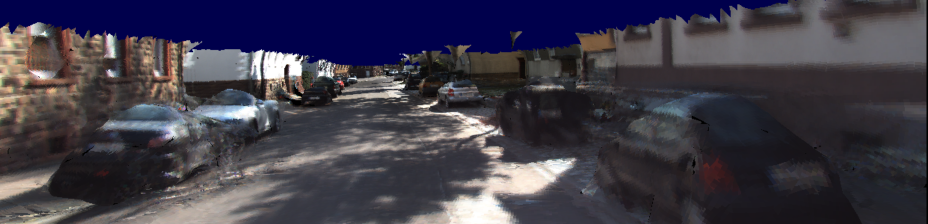}\\
 \end{tabular}
 \caption{Textured mesh reconstructed from the KITTI sequence. Notice how moving objects do not appear in the final results}
 \label{fig:recwithoutmoving}
\end{figure}

In this paper we propose a novel visibility consistent 3D photometric mapping algorithm to reconstruct a urban textured mesh relying on both images and a sparse, although accurate, 3D point clouds, coming from a lidar sensor. The algorithm we propose is robust to the presence of moving object and it is able to reconstruct a consistent textured mesh without them (see in Fig. \ref{fig:recwithoutmoving}).

In Section~\ref{sec:related_work} we discuss works in the literature close to the proposed system. 
In Section~\ref{sec:Textured} we describe the novel framework proposed in the paper, which relies on both lidar data and images; in Section~\ref{sec:experimental} we show the experimental results and in Section~\ref{sec:concl} we conclude the paper proposing out some future direction of research.

\section{Related work}
\label{sec:related_work}
In this paper we present a complete framework to estimate a texturized triangular mesh map of the environment in which moving object are explicitly removed from the geometry of the scene from the photometric refinement process and from the texture.
We now give an overview of the various topic involved in the design of this framework 

Mapping from laser sensors is a well studied research area in Robotics; in the early studies, the map has been estimated in 2 dimensions~\cite{grisettiyz2005improving}, while, in recent years, the prevalent approach is to estimate it in 3D thanks to advances in algorithms, processing and sensors.
Mapping can be pursued together with robot self-localization leading to Simultaneous Localization and Mapping systems; these algorithms do not focus on the mapping part, indeed they reconstruct a sparse point-based map of the environment, while in our case we aim at reconstructing a dense representation of it.

Some approaches estimate a 2.5D map of the environment by populating a grid on the ground plane with the corresponding cell heights~\cite{herbert1989terrain}. These maps are useful for robot navigation, but neglect most of the environment details.
A more coherent representation of the scene is volumetric, i.e,  the space is partitioned into small parts classified as \emph{occupied}, \emph{free} and, in some cases, \emph{unknown}, and the boundary between occupied and free space represents the 3D map. 
In laser-based mapping the most common volumetric representation is voxel-based due to its good trade-off between expressiveness and easiness of implementation~\cite{moravec1996robot};
the drawback of this representation is the large memory consumption, and, therefore its non-scalability.
Many efforts have been directed to improve the scalability and accuracy of voxel based mapping.
Ryde and Hu~\cite{ryde20103d} store only occupied voxels, while  Dryanovski \emph{et al.}~\cite{dryanovski2010multi} store both occupied and free voxels, in order to represent also the uncertainty of unknown space.
The state-of-the-art system OctoMap~\cite{hornung2013octomap}, and its extension~\cite{khan2015adaptive}, are able to efficiently store large maps by including an octree indexing to add flexibility to the framework.

Voxel-based approaches usually produce unappealing reconstructions, due to the voxelization of the space, and they need a very high resolution to capture fine details of the scene, trading off their efficiency.
In Computer Vision community, different volumetric representations have been explored, in particular many algorithms adopt the 3D Delaunay triangulation~\cite{romanoni15a,litvinov_lhiuller14,romanoni15b,vu_et_al_2012}.
Delaunay triangulation is self-adaptive according to the density of the data, i.e., the points, without any indexing policy; moreover its structure is made up of tetraedra from which it is easy to extract a triangular mesh, widely used in the Computer Graphics community to accurately model objects.
These algorithms are consistent with the visibility, i.e., they mark the tetrahedra as free space or occupied according to the camera-to-point rays, assuming that a tetrahedron is empty if one, or at least one, ray intersects them.

\begin{figure}[tp]
 \centering
	\includegraphics[width=0.9\columnwidth]{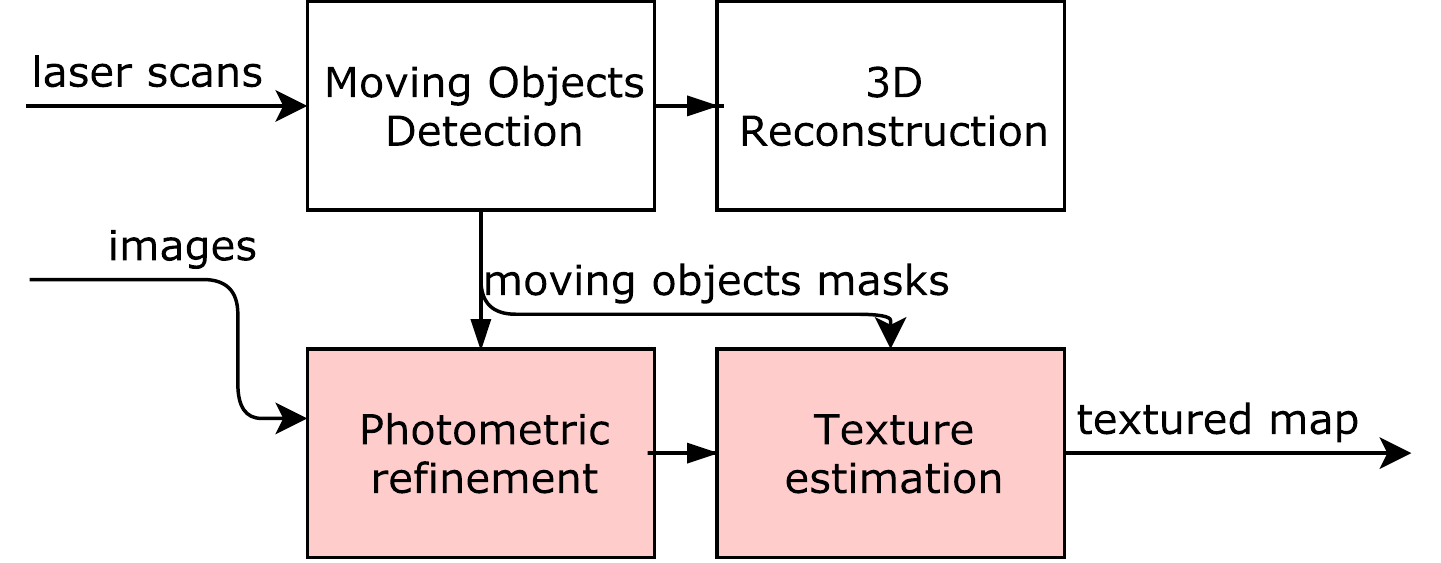}\\
 \caption{Textured mesh pipeline. In this paper we focused on the two red boxes.}
 \label{fig:texturedmesh}
\end{figure}
%In this paper we use laser data to build a first guess for a triangular mesh map leveraging on a Delaunay triangulation, subsequently refined with the photometric image based algorithm. 
Among image-based dense photoconsistent algorithms, the mesh-based algorithm ~\cite{vu_et_al_2012,li2015detail} have been proven to estimate very accurate models and to be scalable in large-scale environments.
They bootstrap form an initial mesh with a volumetric method such as \cite{litvinov_lhiuller14} or \cite{romanoni16} and they refine it by minimizing a photometric energy function defined over the images.
The most relevant drawback happens when moving objects appear in the images: their pixels affect the refinement process leading to inaccurate results.

In our paper, in order to filter out moving objects from the lidar data and the images, we need to explicitly detect them.
A laser-based moving objects detection algorithm has been proposed by Petrovskaya and Thrun \cite{petrovskaya2009model} to detect a moving vehicles using model-based vehicle fitting algorithm; the method performs well, but it needs models for the objects.
%Azim and Aycard~\cite{azim2012detection} look for inconsistencies between subsequent scans: each voxel is classified as free or occupied through ray tracing; voxels classified as both occupied and free in different scans, are called as dynamic. 
%Dynamic voxels are then clustered and filtered such that clusters whose bounding box shape differs significantly from fixed size boxes, are removed. Since boxes are of fixed size, also in this case the  approach is limited to a specific set of objects.
Xiao \emph{et al.}~\cite{xiao2013change} and the Vallet \emph{et al.} \cite{vallet2015extracting} model the physical scanning mechanism of lidar using Dempster-Shafer Theory (DST), evaluating the occupancy of a scan and comparing the consistency among scans. 
A further improvement of these algorithms has been proposed by Postica \emph{et al.} \cite{postica16} where the authors include an image-based validation step which sorts out many false positive.
Pure image-based moving objects detection has been investigated in static camera videos (see \cite{sobral2014comprehensive}), also for the jittering case~\cite{romanoni2014background}, however it is still a very open problem when dealing with moving cameras.

Once moving points have been removed and the photoconsistent map is estimated, a texture can be computed from the images. 
Computer Graphics literature investigated many texuring algorithms \cite{callieri2002reconstructing,alj2012multi,waechter2014let,garcia2013automatic} which however suppose the model to be very accurate. 
In our application we estimate a realistic model useful for navigation purposes, but it still does not capture the fine details of the scene, which are required from these algorithms to work properly.
Especially the resolution is not comparable to the resolution the Computer Graphics algorithm are used to deal with.

\begin{figure*}[tp]
 	\centering
	\setlength{\tabcolsep}{1px}
    \begin{tabular}{cccc}
 	\includegraphics[width=0.24\textwidth]{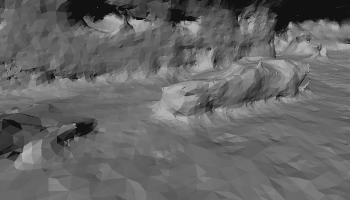}&
	\includegraphics[width=0.24\textwidth]{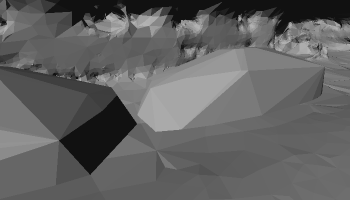}&
	\includegraphics[width=0.24\textwidth]{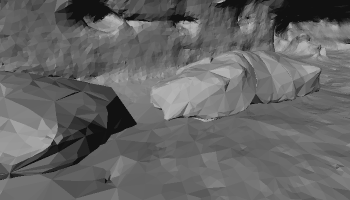}&
	\includegraphics[width=0.24\textwidth]{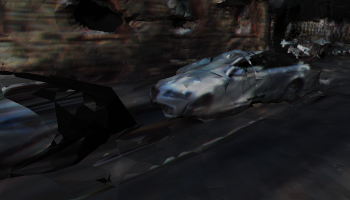}\\
 	(a)&
    (b)&
    (c)&
    (d)
 \end{tabular}
 \caption{Reconstruction of Lidar data: (a) with the visibility consistency~\cite{romanoni15b}; (b) with the car detection; (c) with the  car detection and refinement and (d) after texturing}
 \label{fig:cardetection2}
\end{figure*}

\section{Textured Mesh reconstruction}
\label{sec:Textured}

In Fig. \ref{fig:texturedmesh} we depict the whole pipeline of our system, from the laser and camera data, to the final textured reconstruction.
After an initial preprocessing of laser data, we first detect the moving object, and, from these points, we create a moving object mask corresponding to the images captured by the cameras; then, we estimate a 3D mesh and we refine it leveraging on the information carried by the images and the moving objects masks. 
Finally, we texturize the mesh with a novel efficient algorithm robust with respect  to low resolution meshes.
The camera setting we considered is the monocular one.
The steps of the pipeline are described in the following.

\subsection{Moving Objects Detection and Removal}
\label{subsec:laser}
In the first step of the proposed system we detect the moving objects with the approach proposed in \cite{postica16}. 
Here we briefly describe how it works, but  for an in-depth description we refer the readers to the original paper \cite{postica16}.
We bootstrap from an initial point cloud of the scene, i.e., the first laser scan; as a new scan arrives, we align it to the existing point cloud through the Generalized Iterative Closest Point (GICP)~\cite{segal2009generalized} algorithm and we neglect points too faraway from the sensor, i.e., whose distance from the point cloud center is greater than a given threshold $\tau = 30m$.
To avoid a redundancies we downsample the point cloud by two orders of magnitude; then we estimate and remove ground points. To this extend we model the ground as a grid and we classify points as ground by bootstrapping from the tile occupied by the laser sensor and by applying belief propagation on the grid; the detailed procedure is reported in~\cite{postica16}.

As soon as a set of aligned and filtered scans is available, we detect moving objects as follows. 
Given a point in the space we define a visibility-based rule to estimate its occupancy with respect to a generic laser beam  as $\{empty\}, \{occupied\}, \{unknown\}$, then, for each point $P$ belonging to scan $S_{\text{k}}$ we aggregate evidences for each beam of a scan $S_{\text{i}}$ through Dempster-Shafer Theory (DST) and we classify the point as $\{empty\}$ or $\{occupied\}$.
%by representing the occupancy space using DST.
%The space occupancy is represented using a set $X = \{empty, occupied\}$; DST operates on the power set of $X$, i.e., $2^X = \{\{\emptyset\}, \{empty\}, \{occupied\}, \{empty,occupied\}\}$, where the subset $\{empty,occupied\}$ represents the $unknown$ state, i.e., the space not reached by the beams so far
%DST defines a degree of belief $m(\cdot)$ for each subset: for the empty set it is 0 and for the other subsets they are within the range of $[0, 1]$ and add up to a total of 1.

Then, we classify a point $P$ belonging to a scan $S_{\text{k}}$ as static or moving by comparing its occupancy values with previous and future scans, and computing the probability of a conflicting or consistent state according to DST: if a conflict among states is detected, the corresponding point is classified as moving, otherwise as static.

\subsection{Laser-based 3D Mapping}%%%%%%%%%%%%%%%%%%%%%%%%%%%%%%%%%%%%%%%%%%%%%%%%%%%
%%%%%%%%%%%%%%%%%%%%%%%%%%%%%%%%%%%%%%%%%%%%%%%%%%%%%%%%%%%%%%%%%%%%%%%%%%%%%%%%%%%%
\label{subsec:Textured}
Once the points belonging to moving objects have been removed, we build a map of the environment. 
The 3D laser data and sensor-to-point visibility rays perfectly fits the visibility-consistent approach presented in~\cite{romanoni15b} applied to image-based Structure from Motion (SfM) data; moreover, the algorithm ensures the manifold property of the reconstructed mesh which is needed for the photometric refinement.
However, in real applications, laser beams pass through transparent surfaces, while the SfM points natively adopted by visibility consistent algorithms such as~\cite{litvinov_lhuillier_13} or~\cite{romanoni15b}, do not reconstruct what is behind windows.
In general this does not represent a big issue, but in the presence of cars, laser-based reconstruction is not able to capture adequately the geometry of the scene: rays traverse the interior of the car from different points of view and a visibility consistent reconstruction carves almost all the occupied space, leaving only the lower part of the car in the reconstruction.
To avoid this undesired behavior, we propose to detect the cars in the point cloud and replace their points with a 3D model, e.g., Fig. \ref{fig:cardetection2}.

In our system we implemented a simple hand-crafted car detector being interested here in the joint textured reconstruction from laser and image data.  
We discretize and project the 3D points on the 3D ground plane, then we cluster the set of points which have a rectangular shape. 
Finally, we check if a cluster projected along the longest dimension of the rectangle has a silhouette similar to a car (see the Appendix for a complete explanation).
Learning based approaches could be exploited, for instance, Russel \emph{et al.}~\cite{russell2009associative} or Visin \emph{et al.}~\cite{visin2016reseg}.

% \begin{figure}[tp]
%  \centering
%     \begin{tabular}{cc}
%  	\includegraphics[width=0.22\textwidth]{base00_}&
% 	\includegraphics[width=0.22\textwidth]{car00_}\\
%  	(a)&
%     (b)\\
% 	\includegraphics[width=0.22\textwidth]{nottext00_}&
% 	\includegraphics[width=0.22\textwidth]{text00_}\\
%     (c)&
%     (d)
%  \end{tabular}
%  \caption{Reconstruction of Lidar data: (a) with the visibility consistency~\cite{romanoni15b}; (b) with the car detection; (c) with the  car detection and refinement and (d) after texturing}
%  \label{fig:cardetection2}
% \end{figure}

Once the points belonging to cars have been detected, we remove them from the point cloud of the whole scene, and we group them into a set of clusters:
\begin{equation}
Cars = \left\{c_1, \cdots, c_i, \cdots, c_{N_{cars}}  \right\},
\end{equation}
where $c_i$ represents a single cluster, i.e., a car.

To obtain a consistent reconstruction we also recover the ground points removed for the moving object detection and using the resulting point cloud, the sensor-to-point rays, and the position of the lidar in metric coordinates after each GICP registration, we are able to apply the visibility consistent mesh reconstruction algorithm presented in~\cite{romanoni15b}.
This is a space carving-based method which partitions the space into tetrahedra, and classifies each tetrahedron as free space or matter according to the visibility rays; the boundary between the two classes is the resulting mesh. 

%In order to reconstruct the geometry of the scene we first remove the points belonging to cars from the point cloud estimated during the moving detection stage. The moving object have been yet removed therefore we are able to apply the reconstruction algorithm proposed in \cite{romanoni15b} and \cite{romanoni15b}, where we replace the camera centers with the laser center and the visibility rays goes from laser positions to the observed 3D points.
In this context we do not need to map the environment incrementally as in~\cite{romanoni15b}; therefore, we apply the batch version of the algorithm, i.e., we first add every laser center, 3D point and visibility ray, then we estimate the mesh. 

The 3D reconstruction does not contain the moving points and the cars we removed previously. 
Since we aim at mapping the static part of the scene, we might consider parked cars as a part of it; these cars are then integrated into the final reconstruction. 
Each cluster of points $c_i \in Cars$ represents a car; and to include it into the model of the scene, we first compute its 3D convex hull then we add it to the estimated 3D map.
In Fig. \ref{fig:cardetection2} we show how the car detection affects the final reconstruction.

\begin{figure}[t]
\centering
\includegraphics[width=0.988\columnwidth]{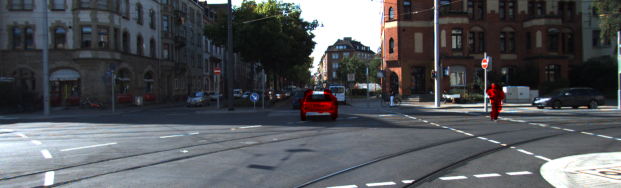}
\caption{Red regions corresponds to the moving objects}
\label{fig:mask}
\end{figure}

\subsection{Photometric Refinement without moving objects}
The removal of points belonging to cars and the explicit inclusion of their convex hull, allows a more consistent reconstruction and it overcomes the issue related to the transparency of the car windows. Now, mapped cars have no more holes, and they are represented more coherently by a convex shape; of course the convex hull is not able to capture all the details, but it is close enough to the real scene to allow for the photometric refinement.

To apply the refinement step, we estimate the mask of the moving objects starting from the detection performed with the algorithm of Postica et al. \cite{postica16}.
To compute the mask corresponding to the $i$-th camera, we project the lidar points detected as moving into the $i$-th image.
Since the resulting mask is very sparse, we first filter it with an all-$1$ 11x11 kernel that grows the detected regions in the neighborhood; then we refine the results by applying a 10px radius disk dilation and a 7px radius disk erosion. In Fig.  \ref{fig:mask} we show an example mask in red overlayed with the original image.

The refinement step described extends the ideas presented in~\cite{vu_et_al_2012} to cope with moving objects. 
The goal is to minimize the energy:
\begin{equation}
\label{eq:en}
E = E_{\textrm{photo}} + E_{\textrm{smooth}} ,
\end{equation}
where  $E_{\textrm{photo}}$ is the data term related to the image photo-consistency measure, and $E_{\textrm{smooth}}$ is a smoothness prior.  

To define the energy $E_{\textrm{photo}}$ let consider two images $I$ and $J$, and the triangular mesh $\mathit{S}$. Let $x$ and $\overrightarrow{n}$ be a point on this mesh and the corresponding normal, and $err_{I, J}(x)$ a function that decreases if the similarity between the patch around the projection of $x$ in $J$ and $I$ increases, then:
\begin{equation}
\label{eq:energy_photo}
  E_{\textrm{photo}} = \sum_{i,j}\int_{\Omega^{\textrm{S}}_{i,j}} err_{I, I_{ij}^{\mathit{S}}}(x_i)\textrm{d}x_i,% = \sum_{i,j} \mathit{err}^{im}_{ij}(x)
\end{equation}
where $I_{ij}^{\mathit{S}}$ is the reprojection of the image from the $j$-th camera in the image $I$ through the mesh $\mathit{S}$ and $\Omega_{i,j}$ represents the domain of the mesh where the projection is defined.
We  minimize Eq. \eqref{eq:energy_photo} through gradient descent by moving each vertex $X_i \in \mathbb{R}^3$ of the mesh  according to the gradient:
\begin{align}
 \begin{split}
  \frac{\textrm{d}E(\mathit{S})}{\textrm{d}X_i} &=  \int_{\mathit{S}} \phi_i(x) \nabla E_{\textrm{photo}}(x) \textrm{d}x,  = \\
  & =  - \sum_{i,j} \int_{\Omega^{\textrm{S}}_{i,j}} \phi_i(x)  f_{ij}(x_i) /(\overrightarrow{n}^T \mathbf{d}_i)\overrightarrow{n} \textrm{d}x_i,
 \end{split}
\end{align}
\begin{equation}
f_{ij}(x_i) =\partial_2 err_{I, I_{ij}^{\mathit{S}}}(x_i) DI_j(x_j) D\Pi_j(x) \mathbf{d}_i,
\end{equation}
where $\phi_i(x)$ represents the barycentric coordinates if $x$ is in the triangle containing $X_i$, otherwise $\phi_i(x) = 0$;
$\Pi_j$ is the j-th camera projection, the vector $\mathbf{d}_i$ goes from camera $i$ to point $x$, the operator $D$ represents the derivative and $\partial_2 err_{I, I_{ij}^{\mathit{S}}}(x_i)$ is the derivative of the similarity measure $err_{ij}(x)$ with respect to the second image.

%  where $\phi_i(x)$ represents the barycentric coordinates if $x$ is in the triangle containing $X_i$, otherwise $\phi_i(x) = 0$, and:
% \begin{equation}
%   \nabla E_{\textrm{photo}}(x) = \nabla (\sum_{i,j} err^{im}_{ij}(x)) = \sum_{i,j} \nabla err^{im}_{ij}(x).
% \end{equation}

% Let $x$ projects in points $x_i = \Pi_i(x) \in I$ and  $x_j = \Pi_j(x) \in I_j$, then, the vector $\mathbf{d}_i$ goes from camera $i$ to point $x$, $z_i$ is the depth of $x$ in camera $i$, and $\overrightarrow{n}$ is the normal at $x$ pointing outward the surface $\mathit{S}$. 
% With the change of variable $\textrm{d}x_i = -\overrightarrow{n}^T \mathbf{d}_i \textrm{d}x/z_i^3$:
% \begin{align}
%  \begin{split}
%   \nabla err^{im}_{ij}(x)& =-\overrightarrow{n} \left( \partial_2 err_{I, I_{ij}^{\mathit{S}}}(x_i) DI_j(x_j) D\Pi_j(x)\frac{\mathbf{d}_i}{z_i^3}\right)=\\
%   &= - f_{ij}(x_i) \overrightarrow{n}/z_i^3.
%  \end{split}
% \end{align}
% where operator $D$ represents the derivative and $\partial_2 err_{I, I_{ij}^{\mathit{S}}}(x_i)$ is the derivative of $err_{ij}(x)$ with respect to the second image.

% We rewrite the discrete gradient as:
% \begin{align}
%  \begin{split}
% \label{eq:final}
%   \frac{\textrm{d}E(\mathit{S})}{\textrm{d}X_i} &=  - \int_{\mathit{S}} \phi_i(x) \sum_{i,j} \nabla err^{int}_{ij}(x) \textrm{d}x \\
% & =  - \sum_{i,j} \int_{\Omega^{\textrm{S}}_{i,j}} \phi_i(x)  f_{ij}(x_i)  \overrightarrow{n}/z_i^3 \frac{z_i^3}{\overrightarrow{n}^T \mathbf{d}_i }\overrightarrow{n} \textrm{d}x_i
%  \end{split}
% \end{align}
We  modify the previous equation in order to discard the moving objects. Let $m_i$ and $m^{\textrm{S}}_{ij}$ the mask of moving objects in the $i$-th camera and the mask of the moving objects in the $j$-th camera projected through $\textrm{S}$ in the camera $i$, we define $\Lambda = \Omega^{\textrm{S}}_{i,j} \cap m_i \cap m^{\textrm{S}}_{ij}$.

\begin{equation}
\frac{\textrm{d}E(\mathit{S})}{\textrm{d}X_i} =  - \sum_{i,j} \int_{\Lambda} \phi_i(x)  f_{ij}(x_i)  \overrightarrow{n}/z_i^3 \frac{z_i^3}{\overrightarrow{n}^T \mathbf{d}_i }\overrightarrow{n} \textrm{d}x_i
\end{equation}.

We minimize the energy $E_{\textrm{smooth}}$ as in \cite{vu_et_al_2012} by means of the Laplace-Beltrami operator approximated with the umbrella operator \cite{wardetzky2007discrete}, which moves each vertex in the mean position of its neighbors.
% During this iterative process we increase the resolution of the mesh through One-to-four midpoint subdivision algorithm until a triangle project in each images with an area smaller than 16 pixels.

\subsection{Mesh texturing}
After the refined map is available, we run the texturing process.
Differently from the existing Computer Graphics texturing algorithms, in the proposed method, we want to be able to estimate incrementally the texture, such that it is possible to color the mesh while the images are acquired by the robot.
Moreover we avoid to texture the map with the moving objects by discarding   the  pixels corresponding to the moving object.

The idea behind our method is to sum the color contributions coming from different images according to a weight given by the perpendicularity of the viewing ray from the camera to the point on the surface.
Given a point $\mathbf{x} \in \textrm{S}$ belonging to the surface of the model,  $\overrightarrow{n}$ the corresponding normal, and the camera center $\mathbf{c}_j$ we define:
\begin{equation}
w_j(\mathbf{x}) = (\overrightarrow{\mathbf{x} -\mathbf{c}_j}) \cdot \overrightarrow{n} = cos\theta
\end{equation},
where $\theta$ is the angle between the normal and the camera to point direction.
Let $c_{k}(\mathbf{x})$ be the color of the point $\mathbf{x}$ at frame $k$, equal to $0$ if $\mathbf{x}$ is not visible; at frame $n+1$ we estimate the color of the texture at location $\mathbf{x}$ as:
\begin{equation}
C_{n+1}(\mathbf{x}) =\frac{W_n(\mathbf{x}) \cdot C_{n}(\mathbf{x}) + w_{n+1}(\mathbf{x})^\alpha \cdot c_{n+1}(\mathbf{x}) }{W_n(\mathbf{x})+ w_{n+1}(\mathbf{x})^\alpha} 
\end{equation}
where
\begin{equation}
W_n(\mathbf{x}) = \sum_{i=1}^n w_i(\mathbf{x})^\alpha, \qquad C_{1}(\mathbf{x}) = c_1(\mathbf{x})
\end{equation}
We raise the weight to the exponent $\alpha$ to increase the importance of weighting, i.e., to increase the importance of the contributions of the pixels which are more  perpendicular to the surface (by experimental evaluation we fixed $\alpha=8$).
% More 

% We texturize the mesh with a simple but effective method, in which we are able to mask the moving object such that the colored map is recovered without them. 

% the Mask Photo Blending algorithm proposed in~\cite{callieri2008masked}, where instead of taking into account all the pixels of all images to texturize the mesh, we project on the mesh only the regions of the images classified as static. 
% At the end of the whole algorithm we aim at visualizing a textured mesh of the scene where moving objects have been removed. 
% For this purpose we need to filter out the image pixels corresponding to moving objects both in the refinement and texturing processes.
% We create a mask of moving objects by expanding the projection of the moving points detected previously through morphological dilation (red regions in Figure \ref{fig:mask}). 

\begin{table}[tp]
\caption{Results on KITTI sequences: map error}
\label{tab:res}
\centering
\setlength{\tabcolsep}{7px}
\begin{tabular}{lcccc}
\toprule           
&\multicolumn{2}{c}{seq 0095}&\multicolumn{2}{c}{seq 0104}\\
&avg & std  &avg & std \\
\midrule
Romanoni \emph{et al.} \cite{romanoni15b} & 0.089&0.131 & 0.194 & 0.311\\
%after car detection  & 0.085&0.099 & 0.087 & 0.123\\
after refinement  & \textbf{0.082}&\textbf{0.098} &  \textbf{0.082}&\textbf{0.103} \\
\end{tabular}
\end{table}

% \begin{table}[t]
% \caption{Dimension of map reconstructed.}
% \label{tab:dimension}
% \centering
% %\setlength{\tabcolsep}{3px}
% \begin{tabular}{lcc}
% \toprule           
% &seq 0095&seq 0104\\
% \midrule
% \cite{hornung2013octomap} & 250 MB&   \\
% \cite{romanoni15b} & 18 MB & 8 MB \\
% Proposed approach  & 28 MB  & 31 MB\\
% \end{tabular}
% \end{table}

\section{Experimental Results}
\label{sec:experimental}
We tested our approach against the publicly available KITTI dataset~\cite{geiger_et_al12}; in particular we used the sequences 0095 and 0104, captured by a Velodyne 64HD with respectively 268 and 313 1392x512 gray scale frames. 
 The algorithm runs on a 4 Core i7-2630QM CPU at 2.2Ghz (6M Cache), with 6GB of DDR3 SDRAM and NVIDIA GeForce GT 630M.
\subsection{Mapping}
To provide a quantitative evaluation of the reconstructed mesh, we compare it  against the full point cloud, i.e., without the downsampling needed for moving object detection (see Section~ \ref{subsec:laser}). Lidar data are dense and accurate enough to be considered as ground truth at least locally. 
We removed from the full point cloud both the moving point and the interior of the cars we do not want to map.
The mesh to point cloud comparison was computed by the tool CloudCompare \cite{cloudcompare} which averages the distances from each point of the ground truth, to the  nearest triangle in the estimated mesh.

We compare our algorithm against different approaches applied to the same downsampled point cloud to provide a fair evaluation: the method proposed in~\cite{romanoni15b}; two widespread algorithms for mesh reconstruction from point clouds (i.e., Poisson Reconstruction~\cite{kazhdan2006poisson} and Ball Pivoting~\cite{bernardini1999ball}); and OctoMap~\cite{hornung2013octomap}, i.e., state-of-the-art laser-based mapping algorithm.

In Table~\ref{tab:res} we show the result of our comparison with~\cite{romanoni15b}. The average errors are below 0.1 m which is enough accurate for a wide variety of robotics tasks, such as localization and navigation. 
The proposed algorithm improves the accuracy of~\cite{romanoni15b}, and both car detection and photometric refinement have contributed to this enhancement; 
since the number of cars in the 0104 sequence is greater than those in sequence 0095, the improvement is more evident in the former case.

\begin{figure}[tp]
 \centering
\setlength{\tabcolsep}{1px}
    \begin{tabular}{c}
	\includegraphics[width=0.48\textwidth]{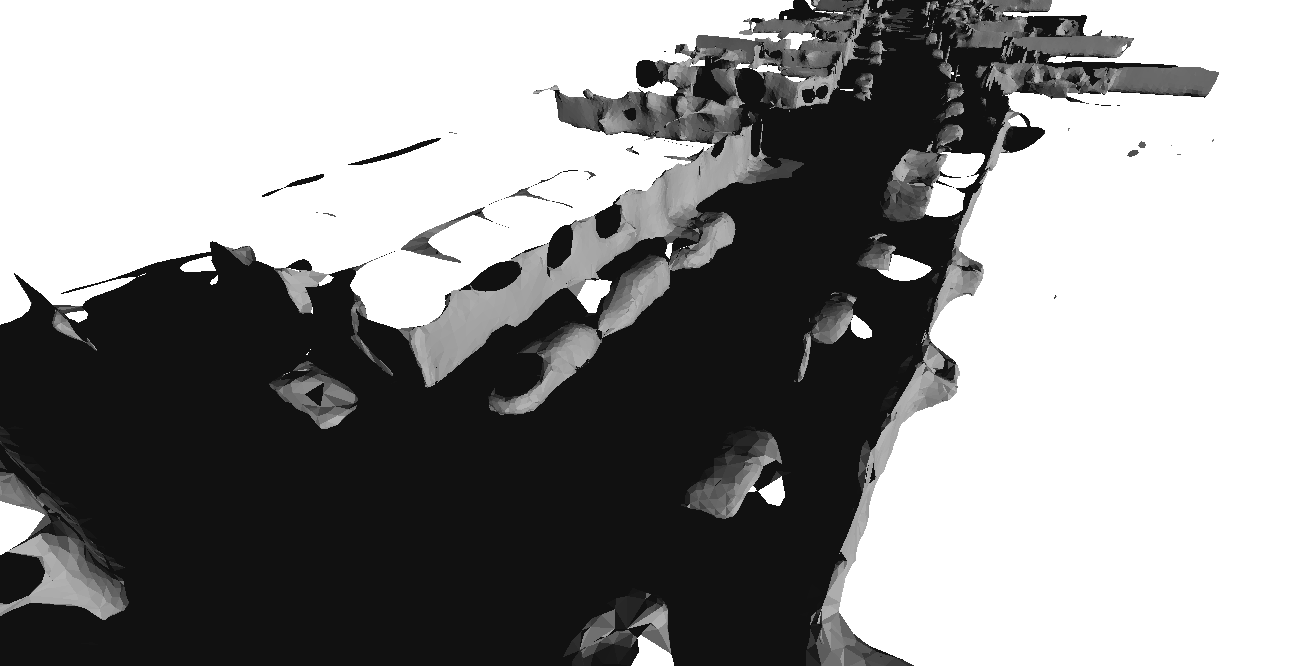}\\
 	\end{tabular}
 \caption{Ball Pivoting reconstruction: dark regions are caused by non consistent facet normals}
 \label{fig:ball}
\end{figure}

\begin{figure*}[tb]
 \centering
\setlength{\tabcolsep}{2px}
    \begin{tabular}{ccc}
 	\includegraphics[width=0.32\textwidth,height=0.16\textwidth]{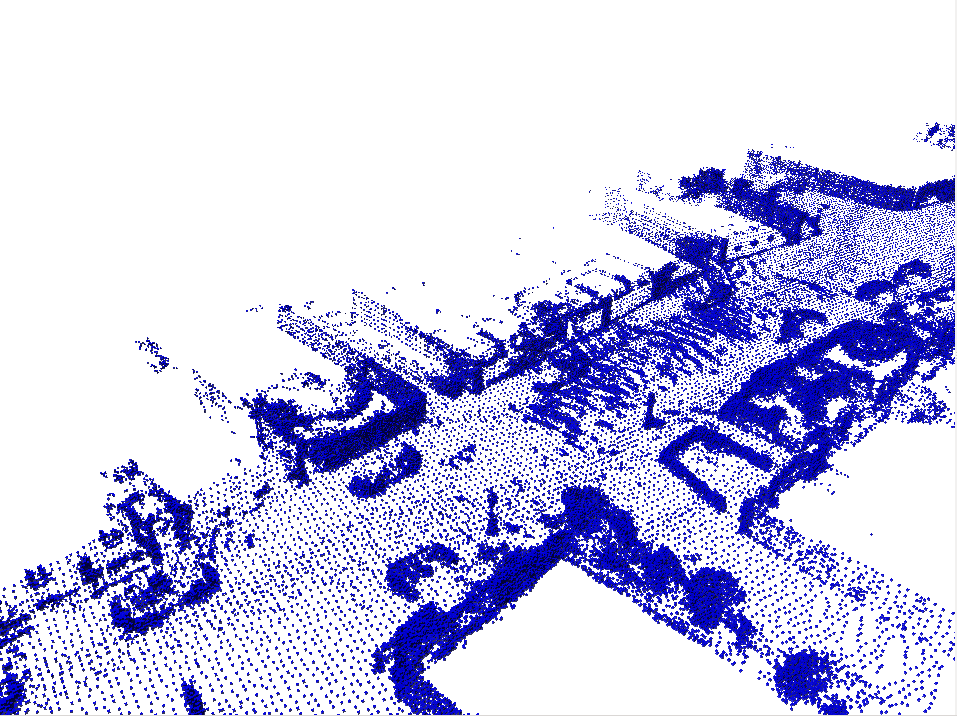}&
	\includegraphics[width=0.32\textwidth,height=0.16\textwidth]{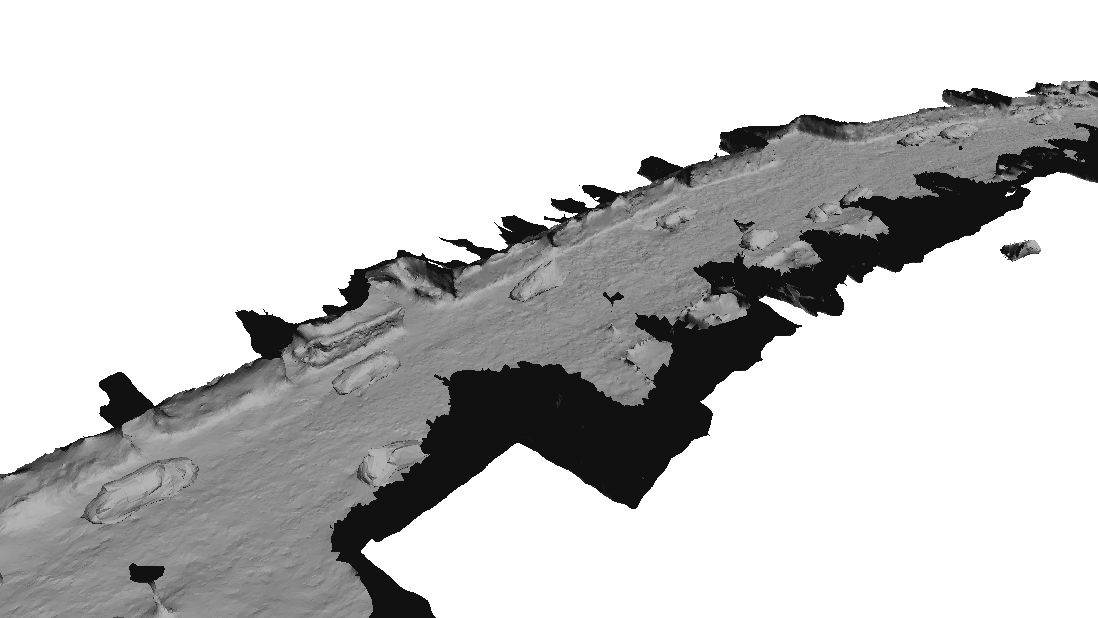}&
	\includegraphics[width=0.32\textwidth,height=0.16\textwidth]{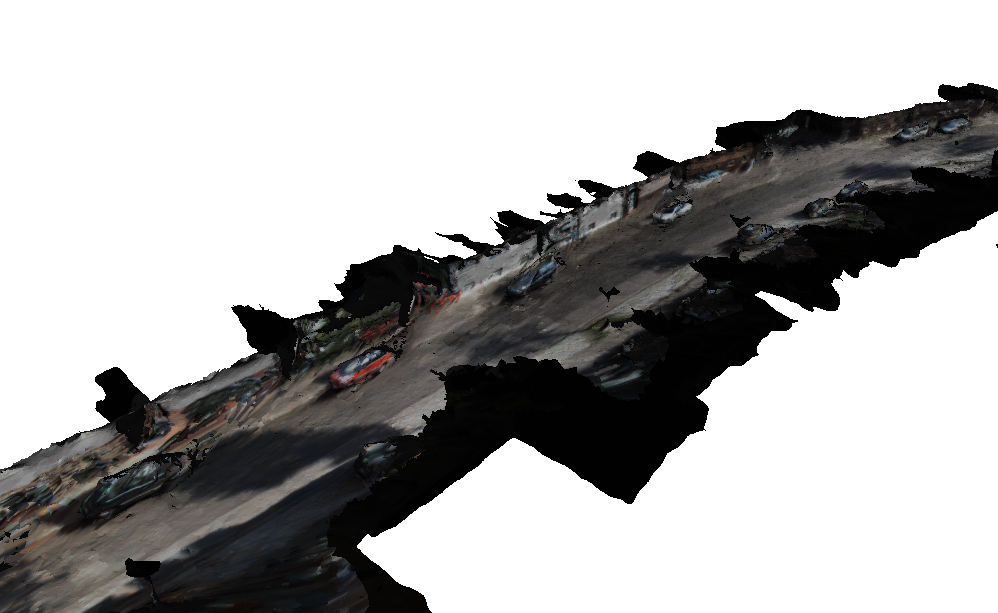}\\
    \multicolumn{3}{c}{Whole 0095 sequence}\\
 	\includegraphics[width=0.32\textwidth,height=0.16\textwidth]{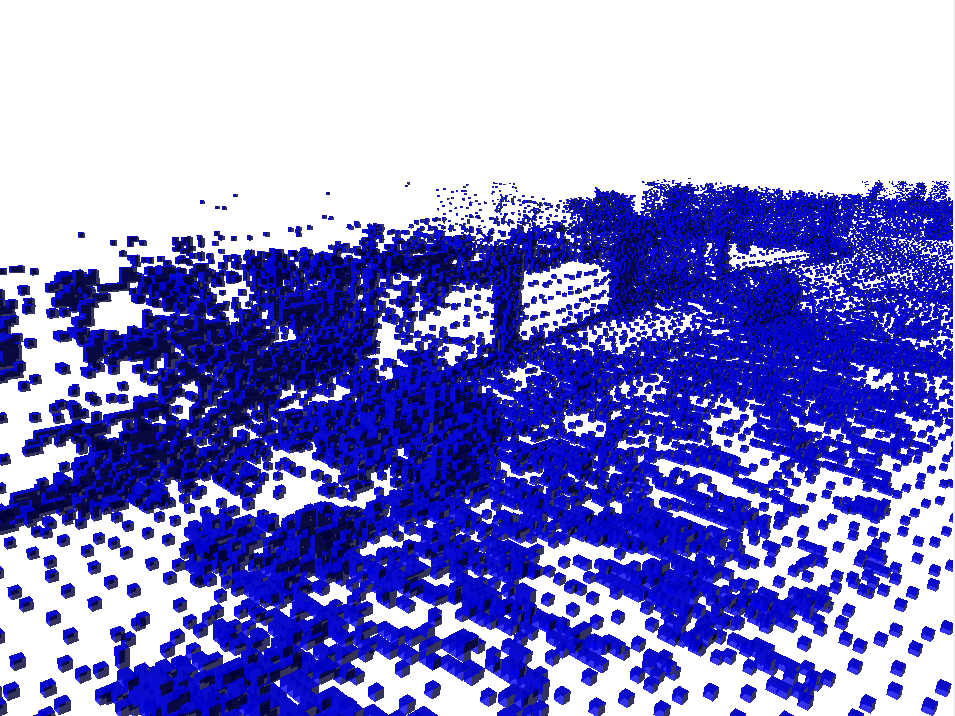}&
	\includegraphics[width=0.32\textwidth,height=0.16\textwidth]{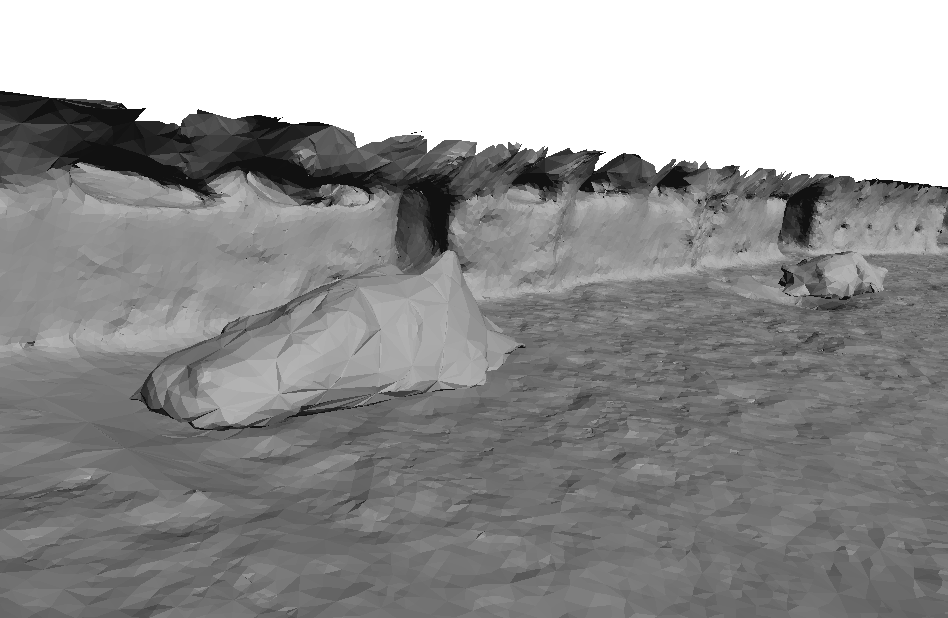}&
	\includegraphics[width=0.32\textwidth,height=0.16\textwidth]{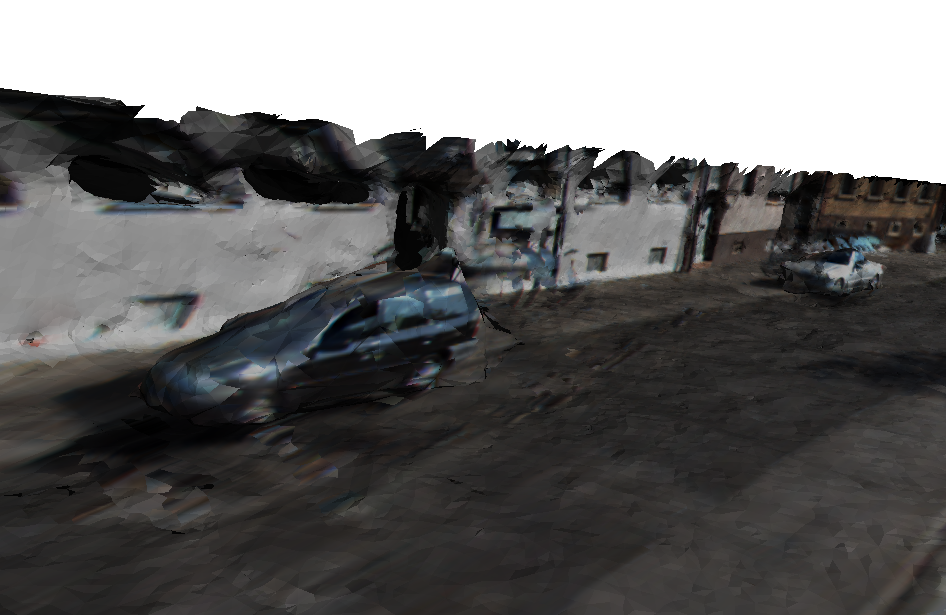}\\
    \multicolumn{3}{c}{Detail of a car}
 \end{tabular}
 \caption{Reconstruction of 0095 sequence with Octomap (first column) and with the proposed approach with and withoud texture (second and third columns)}
 \label{fig:resultOcto01}
\end{figure*}

Poisson Reconstruction was not able to produce a proper reconstruction due to the sparsity of the downsampled laser data; for the same reason Ball Pivoting produces big holes in the reconstruction and Octomap was not able to recover a dense structure. 
In addition to holes, Ball Pivoting reconstructs a mesh whose normals are not consistent and contains severe self intersections (see Fig. \ref{fig:ball}).
Moreover, while the appearance of the results of the proposed algorithm, of~\cite{romanoni15b} and, to some extent, of the Ball Pivoting are realistic, Octomap reconstructs a voxelized map in which many details are lost (see  Fig.~\ref{fig:resultOcto01}).

The algorithm reconstructed and refined the 0095 (268 frames) sequence in 73 minutes and the sequence 0104 (313 frames) in 80 minutes.

Since we rely the reconstruction on laser scans which have been georeferenced, the final map is in turn georeferenced and could be useful for classical robotic localization or navigation tasks. 

\subsection{Texturing} 
\begin{figure}[tb]
 \centering
\setlength{\tabcolsep}{2px}
    \begin{tabular}{c}
	\includegraphics[width=0.45\textwidth]{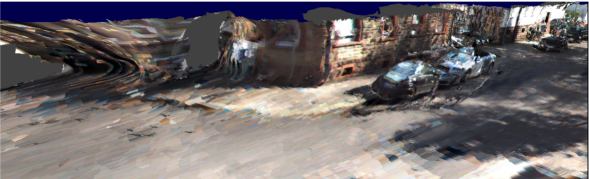}\\
    {Masked Photo Blending \cite{callieri2008masked}}\\
	\includegraphics[width=0.45\textwidth]{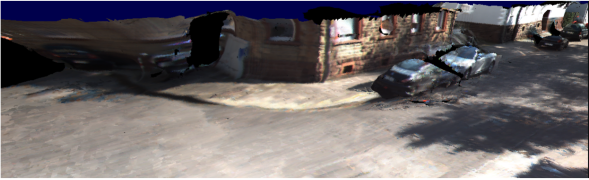}\\
    {Proposed texturing}
 \end{tabular}
 \caption{Texturing results}
 \label{fig:tex01}
\end{figure}

\begin{figure}[tb]
 \centering
\setlength{\tabcolsep}{2px}
    \begin{tabular}{c}
	\includegraphics[width=0.45\textwidth]{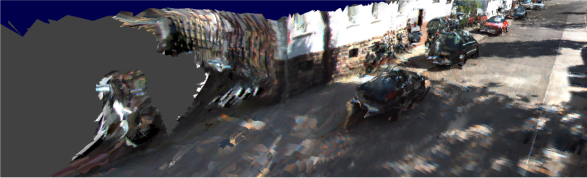}\\
    {Masked Photo Blending \cite{callieri2008masked}}\\
	\includegraphics[width=0.45\textwidth]{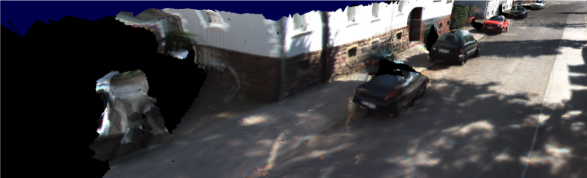}\\
    {Proposed texturing}
 \end{tabular}
 \caption{Texturing results}
 \label{fig:tex02}
\end{figure}

We tested our texturing method against the widely adopted texturing approaches proposed by the Computer Graphic community, i.e., Mask Photo Blending algorithm proposed in~\cite{callieri2008masked}.
Here we evaluate the results by visual inspection since no quantitative evaluation is possible. 
We observed that our method, applied to a typical  Computer Graphics dataset, achieves results comparable with the Mask Photo Blending algorithm, even if it is aimed to texturize lower resolute and less accurate meshes.
In our scenario the proposed texturing algorithm obtains much better results with respect the Mask Photo Blending: it keeps the color continuity among neighboring facets. In Fig. \ref{fig:tex01} and Fig. \ref{fig:tex02} we show two examples of the comparison between the two approaches.

The algorithm texturized the sequence 0095 (268 frames)  in 0.95187 seconds per frame and the sequence 0104 in 1.0374 seconds per frame.
%In Table \ref{tab:dimension} we show that even if our approach is able to capture finer details of the scene, the resulting map is smaller than the map produced by octomap 
%Even if Octomap aims at representing the scene in a compact map, the approach both \cite{romanoni15b} and the proposed approach result in a smaller map.
%confronto tempi e 

% \begin{figure}[tp]
%  \centering
% \setlength{\tabcolsep}{2px}
%     \begin{tabular}{c}
%  	\includegraphics[width=0.48\textwidth]{notCar01}\\
%  	(a) Without cars detection\\
% 	\includegraphics[width=0.48\textwidth]{onlyCar01}\\
%     (c) With cars convex hulls\\
% 	\includegraphics[width=0.48\textwidth]{nottextured01}\\
%     (d) After refinement\\
% 	\includegraphics[width=0.48\textwidth]{tex00}\\
%     (d) Refinement and texturing\\
% 	\includegraphics[width=0.48\textwidth]{0000000016}\\
%     (e) corresponding frame
%  \end{tabular}
%  \caption{Results on frame 16. Let notice that moving objects (the people by bicycle) do not affect the reconstruction}
%  \label{fig:results06}
% \end{figure}

\begin{figure}[tp]
 \centering
\setlength{\tabcolsep}{2px}
    \begin{tabular}{cc}
 	\includegraphics[width=0.24\textwidth,height=0.07\textwidth]{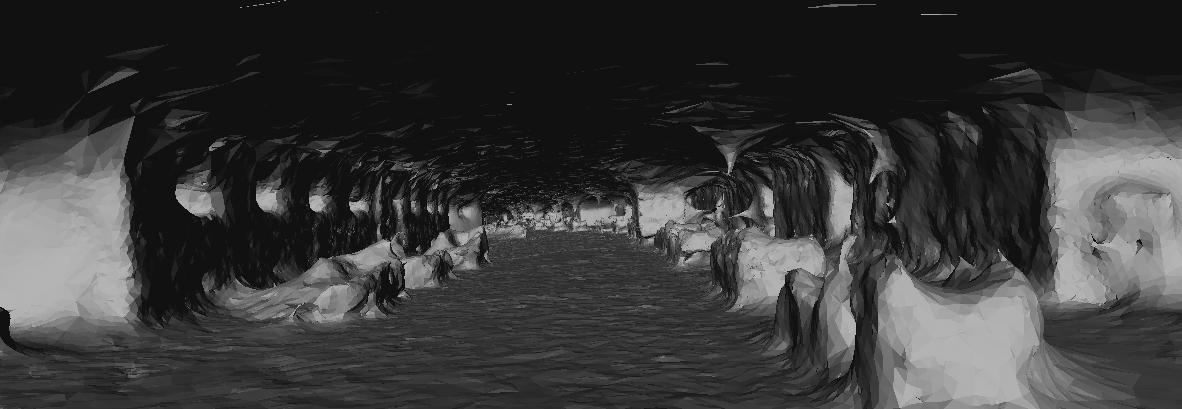}&
	\includegraphics[width=0.24\textwidth,height=0.07\textwidth]{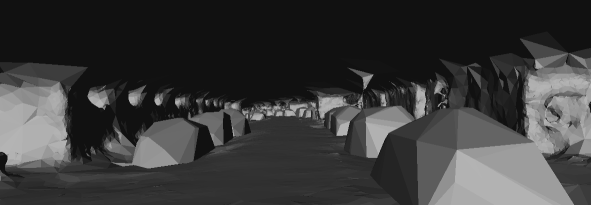}\\
 	(a) Without cars detection&
    (c) With cars convex hulls\\
	\includegraphics[width=0.24\textwidth,height=0.07\textwidth]{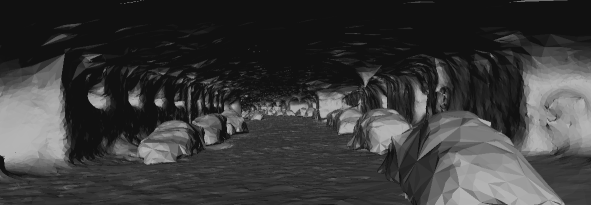}&
	\includegraphics[width=0.24\textwidth,height=0.07\textwidth]{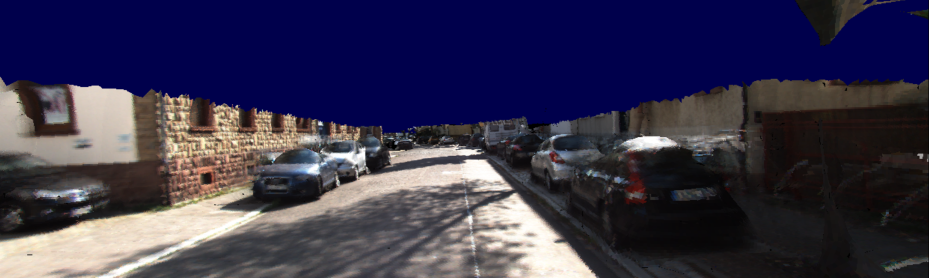}\\
    (d) After refinement&
    (d) Refinement and texturing\\
    \multicolumn{2}{c}{
	\includegraphics[width=0.48\textwidth]{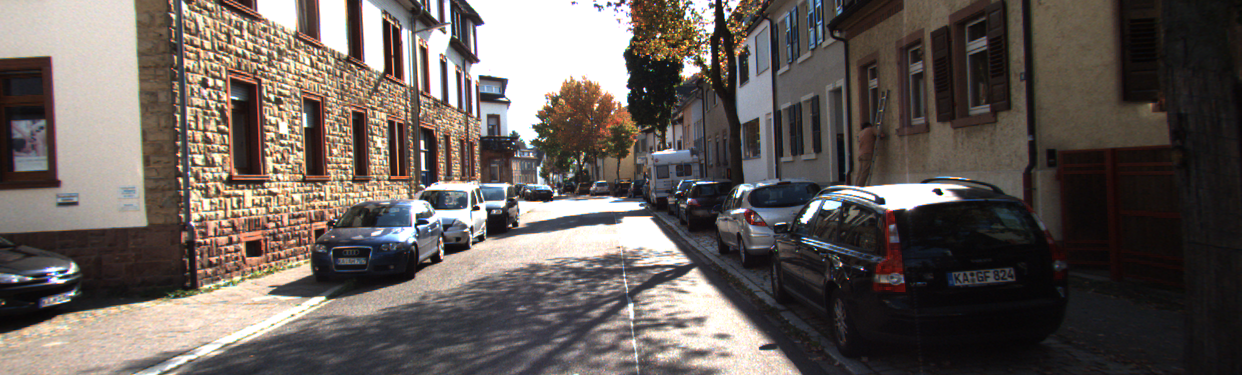}}\\
    \multicolumn{2}{c}{(e) corresponding frame}
 \end{tabular}
 \caption{Results on frame 16. Let notice that moving objects (the people by bicycle) do not affect the reconstruction}
 \label{fig:results06}
\end{figure}

% \begin{figure}[tp]
%  \centering
% \setlength{\tabcolsep}{2px}
%     \begin{tabular}{cc}
%  	\includegraphics[width=0.24\textwidth]{notCar00}&
% 	\includegraphics[width=0.24\textwidth]{car00}\\
%  	(a) Without cars detection&
%     (c) With cars convex hulls\\
% 	\includegraphics[width=0.24\textwidth]{nottextured00}&
% 	\includegraphics[width=0.24\textwidth]{textured00}\\
%     (d) After refinement&
%     (d) Refinement and texturing\\
%     \multicolumn{2}{c}{
% 	\includegraphics[width=0.48\textwidth]{0000000128}}\\
%     \multicolumn{2}{c}{(e) corresponding frame}
%  \end{tabular}
%  \caption{Results on frame 128. Let notice that moving objects (the people by bicycle) do not affect the reconstruction}
%  \label{fig:results}
% \end{figure}

% \begin{figure}[tb]
%  \centering
% \setlength{\tabcolsep}{2px}
%     \begin{tabular}{cc}
% 	\includegraphics[width=0.24\textwidth,height=0.3\textwidth]{dwarf_mesh}&
% 	\includegraphics[width=0.24\textwidth,height=0.3\textwidth]{dwarf_our}\\
%     {Meshlab}&
%     {proposed texturing}
%  \end{tabular}
%  \caption{Texturing results}
%  \label{fig:dwarf}
% \end{figure}

\section{Conclusions and Future Work}
\label{sec:concl}
In this paper we  proposed for the first time a complete framework to extract a 3D textured map of the environment by exploiting both accuracy of the lased data and the dense appearance captured by images. 
We were able to recover a continuous 3D map consistent with the visibility of the laser beams, and refine it by relying on the images.
We explicitly removed moving objects from the laser-data and we detected and modeled independently the cars, which windows are usually traversed  by laser beams.
Finally, we textured the map: both refinement and texturing have not been affected by the presence of moving objects, since we infer their image positions from the laser-based data.

In the future we are interested in improving the convergence of the refinement algorithm adapting the method presented in \cite{romanoni16} and we would exploit the semantics of the scene, e.g., as extracted by deep learning segmentation proposed in~\cite{visin2016reseg}, to further improve the reconstruction.
We could extend the method presented here to recover the cars to multiple class of objects, e.g., trees and poles.
Moreover, we could increase the accuracy of the texturing, with an ad-hoc method which exploits the photometric scores to properly weight the per-triangle texturing.

%\cite{romanoni16}
%\cite{romanoni15b}
%\cite{romanoni2015backward}
%reference wacv e tracking

% \begin{figure}[tp]
%  \centering
%  \includegraphics[width=0.4\textwidth]{drawing.png}
%  \caption{Regions detected as cars}
%  \label{fig:drawing}
% \end{figure}
\section*{Acknowledgments}
{\small
This work has been supported by the POLISOCIAL Grant ``Maps for Easy Paths (MEP)'', the ``Interaction between Driver Road Infrastructure Vehicle and Environment (I.DRIVE)'' Inter-department Laboratory from Politecnico di Milano, and the ``Cloud4Drones'' project founded by EIT Digital. We thank Nvidia who has kindly supported our research through the Hardware Grant Program. }

\bibliographystyle{IEEEtran}
\bibliography{biblio2}

\section*{Appendix}
\label{app}
In this appendix we give the details of the car detection algorithm adopted after the reconstruction step.
First, we project the point cloud on the ground plane and we aggregate the points according to a 2D grid on the $XY$ plane similarly to what we do in the ground removal step; the cell dimension is $0.1$x$0.1$m.
Then we empty the cells containing points higher than a threshold $\tau=2.2m$ to neglect most of the walls and trees, very common in urban environments.
We treat the resulting grid as an image and we apply the closure morphological operator, to close small holes and gaps and to filter out isolated points.
We finally extract connected cells with at least one point and we keep only the regions whose bounding box has a shape compatible with a car, as detected by a template. 
Let $l_{BB}$ and $w_{BB}$ be respectively the length and width of the bounding box around a connected set of cells, and $\rho_{BB}$ the radius circumscribing the box; we filter out all the regions which do not satisfy:
$
 \hat{\rho}_{min} < \rho_{BB} < \hat{\rho}_{max} \quad \text{and} \quad  \hat{r}_{min} < \frac{w_{BB}}{l_{BB}}< \hat{r}_{max},
$
where, in our case, we take into account the average dimensions of the cars and we choose $\hat{\rho}_{min} = 1.5m$, $\hat{\rho}_{max}=5.5m$,  $\hat{r}_{min} =\frac{1.2}{5.0}$ $\hat{r}_{max} = \frac{3.5}{5.0}$.

As a further filtering, we take into account the $Z$ dimension that the previous projection has neglected.
For each 2D bounding box we project the points along the direction parallel to the longest dimension between $l_{BB}$ and $w_{BB}$ in a discrete grid, which is again composed by $0.1m$x$0.1m$ cells.
We compute the convex hull of the projected points and we treat the result as an image. 
For each column we collect the number of white pixels in a vector $\mathbf{\gamma}$; we compute the discrete derivative of $\mathbf{\gamma}$; then, we compute a three bin histogram of it.
The group of points is classified as car if the first and third bin represent respectively an increasing and a decreasing ramps of at least $\frac{\pi}{6}$ and the second bin is almost flat (at most $\frac{\pi}{3}$). 
The method runs in $0.19s$ for a sequence of around $300$m, almost all the cars are detected, moreover, even if some false negative exists, i.e., some cars are not detected, they are going to be reconstructed anyway, at least partially.

\end{document}